\documentstyle[latex-acl,fullname,times]{article}
\newcounter{excerptctr}

\newenvironment{excerpt}[2]
{
 \addtocounter{excerptctr}{1}
 \renewcommand{\excerptinfo}{from utterance #1 in dialog #2}
}{
}

\newenvironment{excerpta}[2]
{\footnotesize
 \begin{quote}
 \renewcommand{\excerptainfo}{(#1 #2)}
}{
  \end{quote}
}

\title{Detecting and Correcting Speech Repairs\thanks{To appear in
`Proceedings of the $32^{nd}$ Annual Meeting of the Association for
Computational Linguistics,' June 1994.}}

\author{Peter Heeman and James Allen \\
Department of Computer Science \\
University of Rochester \\
Rochester, New York, 14627 \\
{\tt \{heeman,james\}@cs.rochester.edu}}

\begin{document}

\bibliographystyle{fullname}

\maketitle{}

\begin{abstract}
Interactive spoken dialog provides many new challenges for spoken
language systems.  One of the most critical is the prevalence of
speech repairs.  This paper presents an algorithm that detects and
corrects speech repairs based on finding the repair pattern.  The
repair pattern is built by finding word matches and word replacements,
and identifying fragments and editing terms.  Rather than using a set
of prebuilt templates, we build the pattern on the fly.  In a fair
test, our method, when combined with a statistical model to filter
possible repairs, was successful at detecting and correcting 80\% of
the repairs, without using prosodic information or a parser.
\end{abstract}

\section{Introduction}

Interactive spoken dialog provides many new challenges for spoken
language systems. One of the most critical is the prevalence of speech
repairs. Speech repairs are dysfluencies where some of the words that
the speaker utters need to be removed in order to correctly understand
the speaker's meaning.  These repairs can be divided into three types:
{\em fresh starts}, {\em modifications}, and {\em abridged}.  A fresh
start is where the speaker abandons what she was saying and starts
again.
\begin{excerpta}{d91-2.2}{utt105}
the current plan is we take -- okay let's say we start with the
bananas \hspace{.2em}(d91-2.2 utt105)
\end{excerpta}
A modification repair is where the speech-repair modifies what was said
before.
\begin{excerpta}{d93-19.3}{utt59}
after the orange juice is at -- the oranges are at the OJ factory
\hspace{.2em}(d93-19.3 utt59)
\end{excerpta}
An abridged repair is where the repair consists solely of a fragment
and/or editing terms.
\begin{excerpta}{d93-14.3}{utt50}
we need to -- um manage to get the bananas to Dansville more quickly
\hspace{.2em} (d93-14.3 utt50)
\end{excerpta}
These examples also illustrate how speech repairs can be divided into three
intervals: the removed text, the editing terms, and the resumed text
(cf.~Levelt, 1983; Nakatani and Hirschberg, 1993).  The removed text, which
might end in a word fragment, is the text that the speaker intends to
replace.  The end of the removed text is called the interruption
point, which is marked in the above examples as ``--''.  This is then
followed by editing terms, which can either be filled pauses, such as
``um'', ``uh'', and ``er'', or cue phrases, such as ``I mean'', ``I
guess'', and ``well''.  The last interval is the resumed text, the
text that is intended to replace the removed text.  (All three
intervals need not be present in a given speech repair.)  In order to
correct a speech repair, the removed text and the editing terms need
to be deleted in order to determine what the speaker intends to
say.\footnote {The removed text and editing terms might still contain
pragmatic information, as the following example displays, ``Peter was
\dots well \dots he was fired.}

In our corpus of problem solving dialogs, 25\% of turns contain at
least one repair, 67\% of repairs occur with at least one other repair
in the turn, and repairs in the same turn occur on average within 6
words of each other.  As a result, no spoken language system will
perform well without an effective way to detect and correct speech
repairs.

We propose that most speech repairs can be detected and
corrected using only local clues---it should not be necessary to test
the syntactic or semantic well-formedness of the entire utterance.
People do not seem to have problems comprehending speech repairs as
they occur, and seem to have no problem even when multiple repairs
occur in the same utterance.  So, it should be possible to construct
an algorithm that runs on-line, processing the input a word at a time,
and committing to whether a string of words is a repair by the end of
the string.  Such an algorithm could precede a parser, or even operate
in lockstep with it.

An ulterior motive for not using higher level syntactic or semantic
knowledge is that the coverage of parsers and semantic interpreters is
not sufficient for unrestricted dialogs.  Recently, \namecite
{dowding-et-al:93:acl} reported syntactic and semantic coverage of
86\% for the DARPA Airline reservation corpus.  Unrestricted dialogs will
present even more difficulties; not only will the speech be less
grammatical, but there is also the problem of segmenting the dialog
into utterance units (cf.~Wang and Hirschberg, 1992).  If speech
repairs can be detected and corrected before parsing and semantic
interpretation, this should simplify those modules as well as make
them more robust.

In this paper, we present an algorithm that detects and corrects
modification and abridged speech repairs without doing syntactic and
semantic processing.  The algorithm determines the text that needs to
be removed by building a repair pattern, based on identification of
word fragments, editing terms, and word correspondences between the
removed and the resumed text (cf.~Bear, Dowding and Shriberg, 1992).
The resulting potential repairs are then passed to a statistical model
that judges the proposal as either fluent speech or an actual repair.

\section{Previous Work}

Several different strategies have been discussed in the literature for
detecting and correcting speech repairs.  A way to compare the
effectiveness of these approaches is to look at their recall and
precision rates.   For detecting repairs, the recall rate is the number
of correctly detected repairs compared to the number of repairs, and
the precision rate is the number of detected repairs compared to the
number of detections (including false positives).  But the true
measures of success are the correction rates.  Correction recall is
the number of repairs that were properly corrected compared to the
number of repairs.  Correction precision is the number of repairs that
were properly corrected compared to the total number of
corrections.

Levelt \shortcite{levelt:83:cog} hypothesized that listeners can use
the following rules for determining the extent of the removed text (he
did not address how a repair could be detected).  If the last word
before the interruption is of the same category as the word before,
then delete the last word before the interruption.  Otherwise, find
the closest word prior to the interruption that is the same as the
first word after the interruption.  That word is the start of the
removed text.  Levelt found that this strategy would work for 50\% of
all repairs (including fresh starts), get 2\% wrong, and have no
comment for the remaining 48\%.\footnote{Levelt claims (pg.~92) that
the hearer can apply his strategy safely for 52\% of all repairs, but
this figure includes the 2\% that the hearer would get wrong.} In
addition, Levelt showed that different editing terms make different
predictions about whether a repair is a fresh start or not.  For
instance, ``uh'' strongly signals an abridged or modification
repair, whereas a word like ``sorry'' signals a fresh start.

\namecite{hindle:83:acl} addressed the problem of correcting
self-repairs by adding rules to a deterministic parser that would
remove the necessary text.  Hindle assumed the presence of an edit
signal that would mark the interruption point, and was able to achieve
a recall rate of 97\% in finding the correct repair.  For modification
repairs, Hindle used three rules for ``expuncting'' text.  The first rule
``is essentially a non-syntactic rule'' that matches repetitions (of
any length); the second matches repeated constituents, both complete;
and the third, matches repeated constituents, in which the first is
not complete, but the second is.

However, Hindle's results are difficult to translate into actual
performance .  First, his parsing strategy depends upon the
``successful disambiguation of the syntactic categories.''  Although
syntactic categories can be determined quite well by their local
context (as is needed by a deterministic parser), Hindle admits that
``[self-repair], by its nature, disrupts the local context.''  Second,
Hindle's algorithm depends on the presence of an edit signal; so far,
however, the abrupt cut-off that some have suggested signals the
repair (cf.~Labov, 1966) has been difficult to find, and it is
unlikely to be represented as a binary feature (cf.~Nakatani and
Hirschberg, 1993).

The SRI group (Bear et al., 1992) employed simple pattern matching
techniques for detecting and correcting modification repairs.\footnote
{They referred to modification repairs as {\em nontrivial} repairs,
and to abridged repairs as {\em trivial} repairs; however, these terms
are misleading.  Consider the utterance ``send it back to Elmira uh to
make OJ''.  Determining that the corrected text should be ``send it
back to Elmira to make OJ'' rather than ``send it back to make OJ'' is
non trivial.} For detection, they were able to achieve a recall rate
of 76\%, and a precision of 62\%, and they were able to find the
correct repair 57\% of the time, leading to an overall correction
recall of 43\% and correction precision of 50\%.  They also tried
combining syntactic and semantic knowledge in a "parser-first"
approach---first try to parse the input and if that fails, invoke
repair strategies based on word patterns in the input.  In a test set
containing 26 repairs \cite {dowding-et-al:93:acl}, they obtained a
detection recall rate of 42\% and a precision of 84.6\%; for
correction, they obtained a recall rate of 30\% and a recall rate of
62\%.

\namecite{nakatani-hirschberg:93:acl}
investigated using acoustic information to detect the interruption
point of speech repairs.  In their corpus, 74\% of all repairs are
marked by a word fragment. Using hand-transcribed prosodic
annotations, they trained a classifier on a 172 utterance training set
to identify the interruption point (each utterance contained at least
one repair).  On a test set of 186 utterances each containing at least
one repair, they obtained a recall rate of 83.4\% and a precision of
93.9\% in detecting speech repairs.  The clues that they found
relevant were duration of pause between words, presence of fragments,
and lexical matching within a window of three words.  However, they do
not address the problem of determining the correction or
distinguishing modification repairs from abridged repairs.

Young and Matessa \cite{young-matessa:91:euro} have also done work in
this area.  In their approach, speech repairs are corrected after a
opportunistic case-frame parser analyzes the utterance.  Their system
looks for parts of the input utterance that were not used by the
parser, and then uses semantic and pragmatic knowledge (of the limited
domain) to correct the interpretation.

\section{The Corpus}
\label{sec:repairmodel}

As part of the {\sc Trains} project \cite{allen-schubert:91:tr}, which
is a long term research project to build a conversationally proficient
planning assistant, we are collecting a corpus of problem solving
dialogs.  The dialogs involve two participants, one who is playing the
role of a user and has a certain task to accomplish, and another, who
is playing the role of the system by acting as a planning
assistant.\footnote{\namecite {gross-allen-traum:92:tr} discuss the
manner in which the first set of dialogues were collected, and provide
transcriptions.} The entire corpus consists of 112 dialogs
totaling almost eight hours in length and containing about 62,000
words, 6300 speaker turns, and 40 different speakers.  These dialogs
have been segmented into utterance files
(cf.~Heeman and Allen, 1994b); words have been transcribed and the
speech repairs have been annotated.  For a training set, we use 40 of
the dialogs, consisting of 24,000 words, 725 modification and abridged
repairs, and 13 speakers; and for testing, 7 of the dialogs,
consisting of 5800 words, 142 modification and abridged repairs, and
seven speakers, none of which were included in the training set.

The speech repairs in the dialog corpus have been hand-annotated.
There is typically a correspondence between the removed text and the
resumed text, and following \namecite {bear-dowding-shriberg:92:acl},
we annotate this using the labels {\bf m} for word matching and {\bf
r} for word replacements (words of the same syntactic category).  Each
pair is given a unique index.  Other words in the removed text and
resumed text are annotated with an {\bf x}.  Also, editing terms
(filled pauses and clue words) are labeled with {\bf et}, and the
moment of interruption with {\bf int}, which will occur before any
editing terms associated with the repair, and after the fragment, if
present.  (Further details of this scheme can be found in
\cite{heeman-allen:94:b}.)  Below is a sample annotation, with removed
text ``go to oran-'', editing term ``um'', and resumed text ``go to''
(d93-14.2 utt60).
\begin{excerpt}{utt60}{d93-14.2}
\begin{verbatim}
go| to|  oran-| um| go| to| Corning
m1| m2| x| int| et| m1| m2|
\end{verbatim}
\end{excerpt}
A speech repair can also be characterized by its {\em repair pattern},
which is a string that consists of the repair labels (word fragments
are labeled as {\bf -}, the interruption point by a period, and
editing terms by {\bf e}).  The repair pattern for the example is {\bf
mm-.emm}.

\section{Repair Indicators}

In order to correct speech repairs, we first need to detect them.  If
we were using prosodic information, we could focus on the actual
interruption point (cf.~Nakatani and Hirschberg, 1993); however, we
are restricting ourselves to lexical clues, and so need to be more
lenient.

Table~\ref{tab:breakdown} gives a breakdown of the modification speech
repairs and the abridged repairs, based on the
hand-annotations.\footnote{Eight repairs were excluded from this
analysis.  These repairs could not be automatically separated from
other repairs that overlapped with them.} Modification repairs are
broken down into four groups, single word repetitions, multiple word
repetitions, one word replacing another, and others.  Also, the
percentage of each type of repair that include fragments and editing
terms is given.
\begin{table}
\begin{center}
\begin{tabular} {|l|c|c|c|} \hline
&       & with    & with Edit \\
& Total & Frag.   &   Term    \\ \hline
Modification Repair   & 450  & 14.7\% & 19.3\% \\
\ \ Word Repetition   & 179  & 16.2\% & 16.2\% \\
\ \ Larger Repetition &  58  & 17.2\% & 19.0\% \\
\ \ Word Replacement  &  72  &  4.2\% & 13.9\% \\
\ \ Other             & 141  & 17.0\% & 26.2\% \\
Abridged Repair       & 267  & 46.4\% & 54.3\% \\ \hline
Total                 & 717  & 26.5\% & 32.4\% \\ \hline
\end{tabular}
\end{center}
\caption{\label{tab:breakdown}Occurrence of Types of Repairs}
\vspace{-.5em}
\end{table}

This table shows that strictly looking for the presence of fragments
and editing terms will miss at least 41\% of speech repairs.  So, we
need to look at word correspondences in order to get better coverage
of our repairs.  In order to keep the false positive rate down, we
restrict ourselves to the following types of word correspondences: (1)
word matching with at most three intervening words, denoted by {\bf
m-m}; (2) two adjacent words matching two others with at most 6 words
intervening, denoted by {\bf mm--mm}; and (3) adjacent replacement,
denoted by {\bf rr}.  Table~\ref{tab:indicators} the number of repairs
in the training corpus that can be deleted by each clue, based on the
hand-annotations.  For each clue, we give the number of repairs that
it will detect in the first column.  In the next three columns, we
give a breakdown of these numbers in terms of how many clues apply.
As the table shows, most repairs are signal by only one of the 3
clues.

\begin{table}[htb]
\begin{center}
\begin{tabular} {|l|c|c|c|c|c|} \hline
& \multicolumn{2}{c|}{Total}      & 1 clue & 2 clues & 3 clues \\ \hline \hline
Fragment      & \multicolumn{2}{c|}{190} &  127 &  58  & 5   \\ \hline
Editing Terms & \multicolumn{2}{c|}{232} &  164 &  63  & 5   \\ \hline
m-m           & 331        &             &      &      &     \\ \cline{1-2}
mm--mm        &  94        &    412      &  296 & 111  & 5   \\ \cline{1-2}
rr            &  59        &             &      &      &     \\ \hline
others        & \multicolumn{2}{c|}  {9} & n.a. & n.a. & n.a.\\ \hline \hline
Total         & \multicolumn{2}{c|}{717} &  587 & 116  & 5   \\ \hline
\end{tabular}
\end{center}
\caption{\label{tab:indicators}Repair Indicators}
\end{table}


Although the {\bf m--m} clue and {\bf mm--mm} clue do not precisely
locate the interruption point, we can, by using simple lexical clues,
detect 97.7\% (708/725) of all the repairs.  But, we still will have a
problem with false positives, and detecting the extent of the repair.

\section{Determining the Correction}

Based on the work done at SRI \cite{bear-dowding-shriberg:92:acl}, we
next looked at the speech repair patterns in our annotated training
corpus.  If we can automatically determine the pattern, then the
deletion of the removed text along with the editing terms gives the
correction.  Since the size of the pattern can be quite large,
especially when editing terms and word fragments are added in, the
number of possible templates becomes very large.  In our training
corpus of 450 modification repairs, we found 72 different patterns
(not including variations due to editing terms and fragments).  All
patterns with at least 2 occurrences are listed in
table~\ref{tab:pat}.
\begin{table}[htb]
\begin{center}
\parbox{1.5in}{\begin{tabular}{|l|r||l|r|} \hline
\makebox[7em][l]{m.m} & \makebox[1.5em][r]{179} \\
r.r & 72 \\
mm.mm & 41 \\
mr.mr & 17 \\
mx.m & 15 \\
mmm.mmm & 14 \\
rm.rm & 12 \\
m.xm & 6 \\
mmr.mmr & 5 \\
m.xxm & 5 \\
x.xx & 4 \\
x. & 4 \\ \hline
\end{tabular}}
\parbox{1.5in}{\begin{tabular}{|l|r||l|r|} \hline
\makebox[7em][l]{mmx.mm} & \makebox[1.5em][r]{4} \\
mrm.mrm & 3 \\
mmmr.mmmr & 3 \\
mm.mxm & 3 \\
r.xr & 2 \\
mxxx.m & 2 \\
mx.mx & 2 \\
mmrm.mmrm & 2 \\
mmmx.mmm & 2 \\
mmmm.mmmm & 2 \\
m.mx & 2 \\
 & \\ \hline
\end{tabular}} \\
\end{center}
\caption{\label{tab:pat}Repair Patterns and Occurrences}
\end{table}

\subsection{Adding to the Pattern}

Rather than doing template matching, we build the repair pattern on
the fly.  When a possible repair is detected, the detection itself
puts constraints on the repair pattern.  For instance, if we detect a
word fragment, the location of the fragment limits the extent of the
editing terms.  It also limits the extent of the resumed text and
removed text, and so on restricts word correspondences that can be
part of the repair.

In this section, we present the rules we use for building repair
patterns.  These rules not only limit the search space, but more
importantly, are intended to keep the number of false positives as
low as possible, by capturing a notion of `well-formness' for speech
repairs.

The four rules listed below follow from the model of repairs that we
presented in the introduction.  They capture how a repair is made up
of three intervals---the removed text, which can end in a word
fragment, possible editing terms, and the resumed text---and how the
interruption point is follows the removed text and precedes the
editing terms.
\begin{quote}
\begin{enumerate}
\item Editing terms must be adjacent.
\item Editing terms must immediately follow the interruption point.
\item A fragment, if present, must immediately precede the interruption point.
\item Word correspondences must straddle the interruption point
and can not be marked on a word labeled as an editing term or
fragment.
\end{enumerate}
\end{quote}

The above rules alone do not restrict the possible word
correspondences enough.  Based on an analysis of the hand-coded
repairs in the training corpus, we propose the following additional
rules.

Rule (5) captures the regularity that word
correspondences of a modification repair are rarely, if ever, embedded
in each other.  Consider the following exception.
\begin{excerpta}{d93-13.1}{utt20}
how would that -- how long that would take
\end{excerpta}
In this example, the word correspondence involving ``that'' is
embedded inside of the correspondence on ``would''.  The speaker
actually made a uncorrected speech error (and so not a speech repair)
in the resumed text, for he should have said ``how long would that
take.''  Without this ungrammaticality, the two correspondences would not have
been embedded, and so would not be in conflict with the following
rule.
\begin{quote}
\begin{enumerate}
\setcounter{enumi}{4}
\item Word correspondences must be cross-serial; a word correspondence
cannot be embedded inside of another correspondence.
\end{enumerate}
\end{quote}

The next rule is used to limit the application of word correspondences
when no correspondences are yet in the repair pattern.  In this case,
the repair would have been detected by the presence of a fragment or
editing terms.  This rule is intended to prevent spurious word
correspondences from being added to the repair.  For instance in the
following example, the correspondence between the two instances of
``I'' is spurious, since the second ``I'' in fact replaces ``we''.
\begin{excerpta}{d93-12.4}{utt76}
I think we need to uh I need
\end{excerpta}
So, when no correspondences are yet included in the repair, the
number of intervening words needs to be limited.  From our test
corpus, we have found that 3 intervening words, excluding fragments
and editing terms is sufficient.
\begin{quote}
\begin{enumerate}
\setcounter{enumi}{5}
\item If there are no other word correspondences, there can only be 3
intervening words, excluding fragments and editing terms, between the
first part and the second part of the correspondence.
\end{enumerate}
\end{quote}

\begin{figure}[htb]
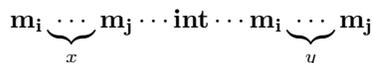

\begin{center}
${\bf m}_{\bf i} \underbrace{\cdots}_x {\bf m}_{\bf j}
\cdots {\bf int} \cdots
{\bf m}_{\bf i} \underbrace{\cdots}_y {\bf m}_{\bf j}$
\vspace{-1em}
\end{center}
\caption{\label{fig:corr}Distance between correspondences}
\end{figure}

The next two rules restrict the distance between two word
correspondences.  Figure~\ref{fig:corr} shows the distance between two
word correspondences, indexed by $i$ and $j$.  The intervals $x$ and
$y$ are sequences of the words that occur between the marked words in
the removed text and in the resumed text, respectively.  The word
correspondences of interest are those that are adjacent, in order
words, the ones that have no labeled words in the $x$ and $y$
intervals.

For two adjacent word correspondences, Rule (7) ensures that there is
at most 4 intervening words in the removed text, and Rule (8) ensures
that there are at most 4 intervening words in the resumed text.
\begin{quote}
\begin{enumerate}
\setcounter{enumi}{6}
\item In the removed text, two adjacent matches can have at most 4 intervening
words ($|x| \le 4$).
\item In the resumed text, two adjacent matches can have at most 4 intervening
words ($|y| \le 4$).
\end{enumerate}
\end{quote}

The next rule, Rule (9), is used to capture the regularity that words
are rarely dropped from the removed text, instead they tend to be
replaced.
\begin{quote}
\begin{enumerate}
\setcounter{enumi}{8}
\item For two adjacent matches, the number of intervening words in the
removed text can be at most one more than the number of intervening
words in the resumed text ($ |x| \le |y| + 1$).
\end{enumerate}
\end{quote}

The last rule, Rule (10), is used to restrict word replacements.  From
an analysis of our corpus, we found that word replacement
correspondences are rarely isolated from other word correspondences.
\begin{quote}
\begin{enumerate}
\setcounter{enumi}{9}
\item A word replacement (except those added by the detection clues)
must either only have fragments and editing terms between the two
words that it marks, or there must be a word correspondence in which
there are no intervening words in either the removed text or the
resumed text ($x = y = 0$).
\end{enumerate}
\end{quote}

\subsection{An Example}

To illustrate the above set of well-formedness constraints on repair
patterns, consider the example given above ``I think we need to -- uh
I need.''  The detection clues will mark the word ``uh'' as being a
possible editing term, giving the partial pattern given below.
\begin{excerpt}{d93-12.4}{utt76}
\begin{verbatim}
I think we need to uh| I need
                   et|
\end{verbatim}
\end{excerpt}

Now let's consider the two instances of ``I''.  Adding this correspondence
to the repair pattern will violate Rule (6), since there are four
intervening words, excluding the editing terms.  The correspondence
between the two instances of `need' is acceptable though, since it
straddles the editing term, and there are only two intervening words
between the corresponding words, excluding editing terms.

Even with the correspondence between the two instances of `need', the
matching between the `I's still cannot be added.  There are 2
intervening words between ``I'' and ``need'' in the removed text, but none
in the resumed side, so this correspondence violates Rule (9).  The
word replacement of ``we'' by the second instance of ``I'', does not
violate any of the rules, including Rule (10), so it is added,
resulting in the following labeling.
\begin{excerpt}{d93-12.4}{utt76}
\begin{verbatim}
I think we| need| to uh| I| need|
         r|    m|    et| r|    m|
\end{verbatim}
\end{excerpt}

\subsection{Algorithm}

Our algorithm for labeling potential repair patterns encodes the assumption
that speech repairs can be
processed one at a time.  The algorithm runs in lockstep with a
part-of-speech tagger \cite{church:88:anlp}, which is used for
deciding possible word replacements.  Words are fed in one at a time.
The detection clues are checked first.  If one of them succeeds, and
there is not a repair being processed, then a new repair pattern is started.
Otherwise, if the clue is consistent with the current repair pattern,
then the pattern is updated; otherwise, the current one is sent off to
be judged, and a new repair pattern is started.

When a new repair is started, a search is made to see if any of the
text can contribute word correspondences to the repair.  Likewise, if
there is currently a repair being built, a search is made to see if
there is a suitable word correspondence for the current word.  Anytime
a correspondence is found, a search is made for any additional
correspondences that it might sanction.

Since there might be a conflict between two possible correspondences
that can be added to a labeling, the one that involves the most recent
pair of words is preferred.  For instance, in the example above, the
correspondence between the second instance of ``I'' and ``we'' is
prefered over the correspondence between the second instance of ``I''
and the first.

The last issue to account for is the judging of a potential repair.
If the labeling consists of just cue phrases, then it is judged as not
being a repair.\footnote{This prevents phrases such as ``I guess''
from being marked as editing terms when they have a sentential
meanings, as in ``I guess we should load the oranges.''} Otherwise, if
the point of interruption of the potential repair is uniquely
determined, then it is taken as a repair.  This will be the case if
there is at least one editing term, a word fragment, or there are no
unaccounted for words between the last removed text part of the last
correspondence and the resumed text part of the first correspondence.

\subsection{Results of Pattern Building}

The input to the algorithm is the word transcriptions, augmented with
turn-taking markers.  Since we are not trying to account for fresh
starts, break points are put in to denote the cancel, and its editing
terms are deleted (this is done to prevent the algorithm from trying
to annotate the fresh start as a repair).  The speech is not marked
with any intonational information, nor is any form of punctuation
inserted.  The results are given in Table~\ref{tab:presults}.
\begin{table}[htb]
\begin{center}
\begin{tabular}{|l|c|c|} \hline
			& Training & Test \\
			&  Set     &  Set \\ \hline
Detection Recall	& 94.9\%   & 91.5\% \\
Detection Precision	& 55.8\%   & 45.3\% \\
Correction Recall	& 89.2\%   & 85.9\% \\
Correction Precision	& 52.4\%   & 42.5\% \\ \hline
\end{tabular}
\end{center}
\caption{\label{tab:presults}Results of Pattern Matching}
\end{table}

The pattern builder gives many false positives in detecting speech
repairs due to word correspondences in fluent speech being
mis-interpreted is evidence of a modification repair.  Also, in
correcting the repairs, word correspondences across an abridged repair
cause the abridged repair to be interpreted as a modification repair,
thus lowering the correction recall rate.\footnote{About half of the
difference between the detection recall rate and the correction recall
rate is due to abridged repairs being misclassified as modification
repairs.} For example, the following abridged repair has two spurious
word correspondences, between ``need to'' and ``manage to''.
\begin{excerpta}{d93-14.3}{utt50}
we need to -- um manage to get the bananas to Dansville more quickly
\end{excerpta}
This spurious word correspondence will cause the pattern builder to
hypothesize that this is a modification repair, and so propose the
wrong correction.

\section{Adding A Statistical Filter}

We make use of a part-of-speech tagger to not only determine
part-of-speech categories (used for deciding possible word
replacements), but also to judge modification repairs that are
proposed by the pattern builder.  For modification repairs, the
category transition probabilities from the last word of the removed
text to the first word of the resumed text have a different
distribution than category transitions for fluent speech.  So, by
giving these distributions to the part-of-speech tagger (obtained from
our test corpus), the tagger can decide if a transition signals a
modification repair or not.

Part-of-speech tagging is the process of assigning to a word the
category that is most probable given the sentential context \cite
{church:88:anlp}. The sentential context is typically approximated by
only a set number of previous categories, usually one or two. Good
part-of-speech results can be obtained using only the preceding
category \cite {weischedel-etal:93:cl}, which is what we will be
using. In this case, the number of states of the Markov model will be
$N$, where $N$ is the number of tags.  By using the Viterbi algorithm,
the part-of-speech tags that lead to the maximum probability path can
be found in linear time.

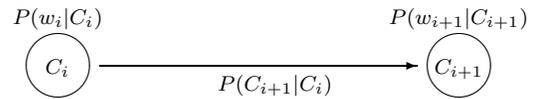
\begin{figure}[htb]
\begin{center}
\setlength{\unitlength}{0.008750in}%
\begingroup\makeatletter\ifx\SetFigFont\undefined
\def\x#1#2#3#4#5#6#7\relax{\def\x{#1#2#3#4#5#6}}%
\expandafter\x\fmtname xxxxxx\relax \def\y{splain}%
\ifx\x\y   
\gdef\SetFigFont#1#2#3{%
  \ifnum #1<17\tiny\else \ifnum #1<20\small\else
  \ifnum #1<24\normalsize\else \ifnum #1<29\large\else
  \ifnum #1<34\Large\else \ifnum #1<41\LARGE\else
     \huge\fi\fi\fi\fi\fi\fi
  \csname #3\endcsname}%
\else
\gdef\SetFigFont#1#2#3{\begingroup
  \count@#1\relax \ifnum 25<\count@\count@25\fi
  \def\x{\endgroup\@setsize\SetFigFont{#2pt}}%
  \expandafter\x
    \csname \romannumeral\the\count@ pt\expandafter\endcsname
    \csname @\romannumeral\the\count@ pt\endcsname
  \csname #3\endcsname}%
\fi
\fi\endgroup
\begin{picture}(332,60)(6,660)
\thinlines
\put( 40,680){\circle{40}}
\put(280,680){\circle{40}}
\put( 65,680){\vector( 1, 0){190}}
\put(170,665){\makebox(0,0)[b]{\smash{\SetFigFont{8}{9.6}{rm}$P(C_{i+1}|C_i )
$}}}
\put( 40,705){\makebox(0,0)[b]{\smash{\SetFigFont{8}{9.6}{rm}$P(w_i|C_i)$}}}
\put( 40,675){\makebox(0,0)[b]{\smash{\SetFigFont{8}{9.6}{rm}$C_i$}}}
\put(280,675){\makebox(0,0)[b]{\smash{\SetFigFont{8}{9.6}{rm}$C_{i+1}$}}}
\put(280,705){\makebox(0,0)[b]{\smash{\SetFigFont{8}{9.6}{rm}$P(w_{i+1}|C_{i+1})$}}}
\end{picture}
\end{center}
\caption{\label{fig:markov1}Markov Model of Part-of-Speech Tagging}
\end{figure}

Figure~\ref{fig:markov1} gives a simplified view of a Markov model for
part-of-speech tagging, where $C_i$ is a possible category for the
$i$th word, $w_i$, and $C_{i+1}$ is a possible category for word
$w_{i+1}$.  The category transition probability is simply the
probability of category $C_{i+1}$ following category $C_i$, which is
written as $P(C_{i+1} | C_i)$.   The probability of word $w_{i+1}$
given category $C_{i+1}$ is $P(w_{i+1}|C_{i+1})$.  The category
assignment that maximizes the product of these probabilities is taken
to be the best category assignment.

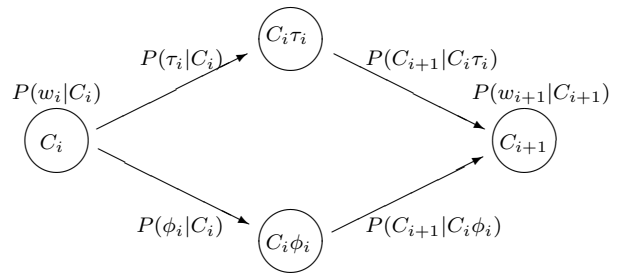
\begin{figure}[htb]
\setlength{\unitlength}{0.008750in}%
\begingroup\makeatletter\ifx\SetFigFont\undefined
\def\x#1#2#3#4#5#6#7\relax{\def\x{#1#2#3#4#5#6}}%
\expandafter\x\fmtname xxxxxx\relax \def\y{splain}%
\ifx\x\y   
\gdef\SetFigFont#1#2#3{%
  \ifnum #1<17\tiny\else \ifnum #1<20\small\else
  \ifnum #1<24\normalsize\else \ifnum #1<29\large\else
  \ifnum #1<34\Large\else \ifnum #1<41\LARGE\else
     \huge\fi\fi\fi\fi\fi\fi
  \csname #3\endcsname}%
\else
\gdef\SetFigFont#1#2#3{\begingroup
  \count@#1\relax \ifnum 25<\count@\count@25\fi
  \def\x{\endgroup\@setsize\SetFigFont{#2pt}}%
  \expandafter\x
    \csname \romannumeral\the\count@ pt\expandafter\endcsname
    \csname @\romannumeral\the\count@ pt\endcsname
  \csname #3\endcsname}%
\fi
\fi\endgroup
\begin{picture}(382,160)(6,600)
\thinlines
\put( 40,680){\circle{40}}
\put(180,620){\circle{40}}
\put(180,741){\circle{38}}
\put(320,680){\circle{40}}
\put( 65,675){\vector( 2,-1){ 90}}
\put( 64,687){\vector( 2, 1){ 90}}
\put(205,625){\vector( 2, 1){ 90}}
\put(206,732){\vector( 2,-1){ 90}}
\put( 30,675){\makebox(0,0)[lb]{\smash{\SetFigFont{8}{9.6}{rm}$C_i$}}}
\put(305,675){\makebox(0,0)[lb]{\smash{\SetFigFont{8}{9.6}{rm}$C_{i+1}$}}}
\put(165,615){\makebox(0,0)[lb]{\smash{\SetFigFont{8}{9.6}{rm}$C_i \phi_i$}}}
\put(165,740){\makebox(0,0)[lb]{\smash{\SetFigFont{8}{9.6}{rm}$C_i  \tau_i$}}}
\put( 40,705){\makebox(0,0)[b]{\smash{\SetFigFont{8}{9.6}{rm}$P(w_i|C_i)$}}}
\put(330,705){\makebox(0,0)[b]{\smash{\SetFigFont{8}{9.6}{rm}$P(w_{i+1}|C_{i+1})$}}}
\put(225,725){\makebox(0,0)[lb]{\smash{\SetFigFont{8}{9.6}{rm}$P(C_{i+1}|C_i
\tau_i) $}}}
\put(140,725){\makebox(0,0)[rb]{\smash{\SetFigFont{8}{9.6}{rm}$P(\tau_i|C_i)$}}}
\put(140,625){\makebox(0,0)[rb]{\smash{\SetFigFont{8}{9.6}{rm}$P(\phi_i|C_i)$}}}
\put(225,625){\makebox(0,0)[lb]{\smash{\SetFigFont{8}{9.6}{rm}$P(C_{i+1}|C_i
\phi_i)$}}}
\end{picture}
\vspace*{-1em}
\caption{\label{fig:simple}Statistical Model of Speech Repairs}
\end{figure}

To incorporate knowledge about modification repairs, we let $R_i$ be a
variable that indicates whether the transition from word $w_i$ to
$w_{i+1}$ contains the interruption point of a modification repair.
Rather than tag each word, $w_i$, with just a category, $C_i$, we will
tag it with $R_{i-1}C_i$, the category and the presence of a
modification repair.  So, we will need the following probabilities,
$P(R_i C_{i+1} | R_{i-1} C_i)$ and $P(w_i | R_{i-1} C_i)$.  To keep
the model simple, and ease problems with sparse data, we make several
independence assumptions.  By assuming that $R_{i-1}$ and $R_i
C_{i+1}$ are independent, given $C_i$, we can simplify the first
probability to $P(R_i | C_i) * P(C_{i+1} | C_i R_i)$; and by assuming
that $R_{i-1}$ and $w_i$ are independent, given $C_i$, we can simplify
the second one to $P(w_i | C_i)$.  The model that results from this is
given in Figure~\ref{fig:simple}.  As can be seen, these manipulations
allow us to view the problem as tagging null tokens between words as
either the interruption point of a modification repair, $R_i =
\tau_i$, or as fluent speech, $R_i = \phi_i$.

Modification repairs can be signaled by other indicators than just
syntactic anomalies.  For instance, word matches, editing terms, and
word fragments also indicate their presence.  This information can be
added in by viewing the presence of such clues as the `word' that is
tagged by the repair indicator $R_i$.  By assuming that these clues
are independent, given the presence of a modification repair, we can
simply use the product of the individual probabilities.  So, the
repair state would have an output probability of $P(F_i|R_i) *
P(E_i|R_i) * P(M_i|R_i)$, where $F_i$, $E_i$, and $M_i$ are random
variables ranging over fragments, editing terms, types of word
matches, respectively.  So for instance, the model can account for how
``uh'' is more likely to signal a modification repair than ``um''.
Further details are given in
\namecite{heeman-allen:94:arpa}.

\section{Overall Results}

The pattern builder on its own gives many false positives due to word
correspondences in fluent speech being mis-interpreted evidence of a
modification repair, and due to word correspondences across an
abridged repair causing the abridged repair to be interpreted as a
modification repair.  This results in an overall correction recall
rate of 86\% and a precision rate of 43\%.  However, the real result
comes from coupling the pattern builder with the decision routine,
which will eliminate most of the false positives.

Potential repairs are divided into two groups.  The first includes
abridged repairs and modification repairs involving only word
repetitions.  These are classified as repairs outright.  The rest of
the modification repairs are judged by the statistical model.  Any
potential repair that it rejects, but which contains a word fragment
or filled pause is accepted as an abridged repair.
Table~\ref{tab:finalresults} gives the results of the combined
approach on the training and test sets.
\begin{table}[htb]
\begin{center}\small
\begin{tabular}{|l|c|c|c|c|c|} \hline
            & Training  & Test    \\
            & Corpus    & Corpus  \\ \hline
Detection   &         &     \\
\ Recall    & 91\%  & 83\%  \\
\ Precision & 96\%  & 89\%  \\ \hline
Correction  &       &       \\
\ Recall    & 88\%  & 80\%  \\
\ Precision & 93\%  & 86\%  \\ \hline
\end{tabular}
\caption{\label{tab:finalresults}Overall Results}
\end{center}
\vspace{-.5em}
\end{table}

Comparing our results to others that have been reported in the
literature must be done with caution. Such a comparison is limited due
to differences in both the type of repairs that are being studied and
in the datasets used for drawing results.  Bear, Dowding, and Shriberg
(1992) use the ATIS corpus, which is a collection of queries made to
an automated airline reservation system.  As stated earlier, they
removed all utterances that contained abridged repairs.  For detection
they obtained a recall rate of 76\% and a precision of 62\%, and for
correction, a recall rate of 43\% and a precision of 50\%.  It is not
clear whether their results would be better or worse if abridged
repairs were included.  \namecite{dowding-et-al:93:acl} used a similar
setup for their data.  As part of a complete system, they obtained a
detection recall rate of 42\% and a precision of 85\%; and for
correction, a recall rate of 30\% and a precision of 62\%.  Lastly,
\namecite {nakatani-hirschberg:93:acl} also used the ATIS corpus, but
in this case, focused only on detection, but detection of all three
types of repairs.  However, their test corpus consisted entirely of
utterances that contained at least one repair.  This makes it hard to
evaluate their results, reporting a detection recall rate of 83\% and
precision of 94\%.  Testing on an entire corpus would clearly decrease
their precision.  As for our own data, we used a corpus of natural
dialogues that were segmented only by speaker turns, not by individual
utterances, and we focused on modification repairs and abridged
repairs, with fresh starts being marked in the input so as not to
cause interference in detecting the other two types.

The performance of our algorithm for correction is significantly
better than other previously reported work, with a recall rate of
80.2\% and a precision rate of 86.4\% on a fair test.  While Nakatani
and Hirschberg report comparable detection rates, and Hindle reports
better correction rates, neither of these researchers attack the
complete problem of both detection and correction. Both of them also
depend on externally supplied annotations not automatically derived
from the input.  As for the SRI work, their parser-first strategy and
simple repair patterns cause their rates to be much lower than ours.
A lot of speech repairs do not look ill-formed, such as ``and a boxcar
of -- and a tanker of OJ'', and ``and bring -- and then bring that
orange juice,'' and are mainly signaled by either lexical or acoustic
clues.

\section{Overlapping Repairs}

Our algorithm is also novel in that it handles overlapping repairs.
Two repairs overlap if part of the text is used in both repairs.
Such repairs occur fairly frequently in our corpus, and for the most
part, our method of processing repairs, even overlapping ones, in a
sequential fashion appears successful.  Out of the 725 modification
and abridged repairs in the training corpus, 164 of them are
overlapping repairs, and our algorithm is able to detect and correct
86.6\% of them, which is just slightly less than the correction recall
rate for all modification and abridged repairs in the entire training corpus.

Consider the following example (d93-14.2 utt26), which contains four
speech repairs, with the last one overlapping the first three.
\begin{excerpta}{d93-14.2}{utt26}
and pick up um the en- I guess the entire um p- pick up the load of
oranges at Corning
\end{excerpta}
The algorithm is fed one word at a time.  When it encounters the first
``um'', the detection rule for editing terms gets activated, and so a
repair pattern is started, with ``um'' being labeled as an editing
term.  The algorithm then processes the word ``the'', for which it can
find no suitable correspondences.  Next is the fragment ``en-''.  This
causes the detection rule for fragments to fire. Since this fragment
comes after the editing term in the repair being built, adding
it to the  repair would violate Rule (2) and Rule (3).  So, the
algorithm must finish with the current repair, the one involving
``um''.  Since this consists of just a filled pause, it is judged as
being an actual repair.

Now that the algorithm is finished with the repair involving ``um'',
it can move on to the next one, the one signaled by the fragment
``en-''.  The next words that are encountered are ``I guess'', which
get labeled as an editing phrase.  The next token is the word ``the'',
for which the algorithm finds a word correspondence with the previous
instance of ``the''.  At this point, it realizes that the repair is
complete (since there is a word correspondence and all words between
the first marked word and the last are accounted for) and so sends it
off to be judged by the statistical model.  The model tags it as a
repair.  Deleting the removed text and the editing terms indicated by
the labeling results in the following, with the algorithm currently
processing ``the''.
\begin{quote}\small
and pick up the entire um p- pick up the load of oranges at Corning
\end{quote}

Continuing on, the next potential repair is triggered by the presence
of ``um'', which is labeled as an editing term.  The next token
encountered, a fragment, also indicates a potential repair, but adding
it to the labeling will violate Rule (2) and Rule (3).  So, the
pattern builder is forced to finish up with the potential repair
involving ``um''.  Since this consists of just a filled pause, it is
accepted.  This leaves us with the following text, with the algorithm
currently processing ``p-'', which it has marked as a fragment.
\begin{quote}\small
and pick up the entire p- pick up the load of oranges at Corning
\end{quote}
The next word it encounters is ``pick''.  This word is too far from
the preceding ``pick'' to allow this correspondence to be added.
However, the detection clue {\bf mm--mm} does fire, due to the
matching of the pair of
adjacent words ``pick up''.  This clue is consistent with
``p-'' being marked as the word fragment of the repair, and so these
correspondences are added.  The next token encountered is ``the'', and
the correspondence for it is found.  Then ``load'' is processed, but
no correspondence is found for it, nor for the remaining words.  So,
the repair pattern that is built contains an unlabeled token, namely
``entire''.  But due to the presence of the word fragment, the
interruption point can be determined.  The repair pattern is set off
to be judged, which tags it as a repair.  This leaves the following
text not labeled as the removed text nor as the editing terms of a repair.
\begin{quote}\small
and pick up the load of oranges at Corning
\end{quote}
Due to the sequential processing of the algorithm and its ability to
commit to a repair without seeing the entire utterance, overlapping
repairs do not pose a major problem.

Some overlapping repairs can cause problems however.  Problems can
occur when word correspondences are attributed to the wrong repair.
Consider the following example (d93-15.2 utt46).
\begin{excerpta}{d93-15.2}{utt46}
you have w- one you have two boxcar
\end{excerpta}
This utterance contains two speech repairs, the first is the
replacement of ``w-'' by ``one'', and the second the replacement of
``you have one'' by ``you have two''.  Since no analysis of fragments
is done, the correspondence between ``w-'' and ``one'' is not
detected.  So, our greedy algorithm decides that the repair after
``w-'' also contains the word matches for ``you'' and ``have'', and
that the occurrence of ``one'' after the ``w-'' is an inserted word.
Due to the presence of the partial and the word matching, the
statistical model accepts this proposal, which leads to the erroneous
correction of ``one you have two boxcars,''  which blocks the
subsequent repair from being found.

\section{Conclusion}

This paper described a method of locally detecting and correction
modification and abridged speech repairs.  Our work shows that a large
percentage of speech repairs can be resolved prior to parsing.
Our algorithm assumes that the speech recognizer produces a sequence
of words and identifies the presence of word fragments.  With the
exception of identifying fresh starts, all other processing is
automatic and does not require additional hand-tailored transcription.
We will be incorporating this method of detecting and correcting
speech repairs into the next version of the {\sc Trains} system, which
will use spoken input.

There is an interesting question as to how good the performance can
get before a parser is required in the process.  Clearly, some
examples require a parser. For instance, we can not account for the
replacement of a noun phrase with a pronoun, as in ``the engine can
take as many um -- it can take up to three loaded boxcars'' without
using syntactic knowledge.  On the other hand, we can expect to improve
on our performance significantly before requiring a parser. The scores
on the training set, as indicated in table~\ref{tab:finalresults},
suggest that we do not have enough training data yet.  In addition, we
do not yet use any prosodic cues.  We are currently investigating
methods of automatically extracting simple prosodic measures that can
be incorporated into the algorithm. Given Nakatani and Hirschberg's
results, there is reason to believe that this would significantly
improve our performance.

Although we did not address fresh starts, we feel that our approach of
combining local information from editing terms, word fragments, and
syntactic anomalies will be successful in detecting them.  However,
the problem lies in determining the extent of the removed text.  In
our corpus of spoken dialogues, the speaker might make several
contributions in a turn, and without incorporating other knowledge, it
is difficult to determine the extent of the text that needs to be
removed.  We are currently investigating approaches to automatically
segment a turn into separate utterance units by using prosodic
information.

\section*{Acknowledgments}

We wish to thank Bin Li, Greg Mitchell, and Mia Stern for their help
in both transcribing and giving us useful comments on the annotation
scheme.  We also wish to thank Hannah Blau, John Dowding, Elizabeth
Shriberg, and David Traum for helpful comments.  Funding gratefully
received from the Natural Sciences and Engineering Research Council of
Canada, from NSF under Grant IRI-90-13160, and from ONR/DARPA under
Grant N00014-92-J-1512.

{\small

}
\end{document}